\pdfoutput=1

\documentclass[11pt]{article}

\usepackage{booktabs}

\usepackage{diagbox}
\usepackage{graphicx}
\usepackage{rotating}      

\usepackage[]{acl}

\usepackage{microtype}

\usepackage[acronym]{glossaries}
\glsdisablehyper

\newacronym{mlp}{MLP}{Multi-Layer Perceptron}
\newacronym{am}{AM}{Argument Mining}

\newacronym{STSD}{Single Task Single Dataset}{Single Task Single Dataset}
\newacronym{STLOOD}{Single Task Leave One Out Dataset}{Single Task Leave One Out Dataset}
\newacronym{STAD}{Single Task All-in Dataset}{Single Task All-in Dataset}
\newacronym{MT}{Multi Task}{Multi Task}
\newacronym{MTLOOD}{Multi Task Leave One Out Dataset}{Single Task Leave One Out Dataset}

\newacronym{Alm}{Alm}{Alm}
\newacronym{ISEAR}{ISEAR}{ISEAR}
\newacronym{SemEval-2007}{SemEval-2007}{SemEval-2007}
\newacronym{SemEval-2018}{SemEval-2018}{SemEval-2018}
\newacronym{SemEval-2019}{SemEval-2019}{SemEval-2019}
\newacronym{Neviarouskaya 2010}{Neviarouskaya 2010}{Neviarouskaya 2010}
\newacronym{Neviarouskaya 2011}{Neviarouskaya 2011}{Neviarouskaya 2011}

\newacronym{IBM-Rank-30k}{IBM-Rank}{IBM-Rank}
\newacronym{IBM-ArgQ}{IBM-ArgQ}{IBM-ArgQ}
\newacronym{SwanRank}{SwanRank}{SwanRank}
\newacronym{UKPConvArgRank}{UKPConvArg}{UKPConvArg}

\newacronym{IBM-Evidence}{IBM-Evidence}{IBM-Evidence}
\newacronym{UKP-Sentential}{UKP-Sentential}{UKP-Sentential}

\newacronym{Argument Identification}{AId}{Argument Identification}
\newacronym{Evidence Detection}{ED}{Evidence Detection}
\newacronym{Argument Quality}{AQ}{Argument Quality}

\newacronym{labeledArgEmotions}{EmoArg-523}{EmoArg-523}

\newacronym{BaseMaj}{Majority Baseline}{Majority Baseline}
\newacronym{BasePro}{Pronouns Baseline}{Pronouns Baseline}
\newacronym{BaseNRC}{NRC Baseline}{NRC Baseline}

\newacronym{HP}{Human Performance}{Human Performance}
\newacronym{BertBaseMT}{EmoBERT}{EmoBERT}
\newacronym{BertBaseMTAM}{ArgBERT-EmoInit}{ArgBERT-EmoInit}
\newacronym{BertBaseAM}{ArgBERT}{ArgBERT}
\newacronym{BertLargeAM}{ArgBERT-L}{ArgBERT-L}
\newacronym{ToBert}{ToBERT}{ToBERT}

\newacronym{EmoArg523}{EmoArg-523}{EmoArg-523}

\usepackage{hyperref}

\usepackage[capitalize]{cleveref}

\usepackage{tabularx}

\usepackage{tabularx}
\usepackage{multirow}

\title{Towards a Holistic View on Argument Quality Prediction}

\author{
    Michael Fromm,\textsuperscript{\rm 1}
    Max Berrendorf,\textsuperscript{\rm 1}
    Johanna Reiml,\textsuperscript{\rm 2} 
    Isabelle Mayerhofer,\textsuperscript{\rm 2}\\
    \bf{Siddharth Bhargava},\textsuperscript{\rm 2}
    \bf{Evgeniy Faerman}\textsuperscript{\rm 1}
    \normalfont{and} \bf{Thomas Seidl\textsuperscript{\rm 1}}\\
    \textsuperscript{\rm 1} \small Database Systems and Data Mining, LMU Munich, Germany \\
    \textsuperscript{\rm 2} \small LMU Munich, Germany \\
    \small \{fromm, berrendorf\}@dbs.ifi.lmu.de
}

\begin{document}
\maketitle
\begin{abstract}
Argumentation is one of society's foundational pillars, and, sparked by advances in NLP and the vast availability of text data, automated mining of arguments receives increasing attention.
A decisive property of arguments is their strength or quality. While there are works on the automated estimation of argument strength, their scope is narrow:
they focus on isolated datasets and neglect the interactions with related argument mining tasks, such as argument identification, evidence detection, or emotional appeal.
In this work, we close this gap by approaching argument quality estimation from multiple different angles:
Grounded on rich results from thorough empirical evaluations, we assess the generalization capabilities of argument quality estimation across diverse domains, the interplay with related argument mining tasks, and the impact of emotions on perceived argument strength.
We find that generalization depends on a sufficient representation of different domains in the training part.
In zero-shot transfer and multi-task experiments, we reveal that argument quality is among the more challenging tasks but can improve others.
Finally, we show that emotions play a minor role in argument quality than is often assumed.
We publish our code at \url{https://anonymous.4open.science/r/kdd-holistic-view-aq-C0D8}.
\end{abstract}

\section{Introduction}
\glsresetall 
The argumentation process is one of the cornerstones of society, as it allows the exchange of opinions and reaching a consensus together.
Fueled by advances in natural language processing, recent years have witnessed the advent of \gls{am}, i.e., the field of automated discovery and organization of arguments.
\gls{am} is helpful over various scenarios, reaching from legal reasoning~\cite{WynerPMM10, walker-etal-2014-annotating, poudyal-etal-2020-echr, villata2020using} to supporting the decision-making process of politicians~\cite{Lippi_Torroni_2016, haddadan-etal-2019-yes, 0eac738616614094950bb74635ce3d49, menini-etal-2017-topic, lippi2016argument, 10.1145/2124295.2124359}.
Thus, there is a flurry of works on identification of arguments from text~\cite{stab-etal-2018-cross, DBLP:conf/webi/FrommF019, trautmann2020fine} and retrieval of them~\cite{wachsmuth2017building, fromm2020diversity, dumani2019systematic, dumani2020framework, stab2018argumentext}.
Since arguments often have to be weighed against each other, a central property of arguments is their \gls{Argument Quality}  or \emph{convincingness}, i.e., their (perceived) strength.
While the ancient Greeks~\cite{2002rhetorik} already discussed the constituents of strong arguments, automated estimation is a relatively uncharted field.
Due to the high subjectivity of argument strength~\cite{swanson-etal-2015-argument, Gretz_Friedman_Cohen-Karlik_Toledo_Lahav_Aharonov_Slonim_2020, toledo2019automatic,habernal-gurevych-2016-argument, stab-etal-2018-cross}, obtaining high-quality annotations is challenging. 
In this light, a legitimate question is the reliability and robustness of the existing approaches for estimating \gls{Argument Quality} and their applicability in real-life scenarios.
Existing \gls{Argument Quality} benchmark datasets are often restricted to a single domain~\cite{wachsmuth2016using, persing2017lightly} or/and make different assumptions about  factors impacting the \gls{Argument Quality}.
Thus, enabling transfer between sources and datasets appears especially appealing, but existing works~\cite{Gretz_Friedman_Cohen-Karlik_Toledo_Lahav_Aharonov_Slonim_2020,toledo2019automatic,swanson-etal-2015-argument, habernal-gurevych-2016-argument} cease to provide detailed studies thereupon.
%
Moreover, social science research suggests that the strength of an argument depends less on its logical coherence and depends much more appealing to the recipient's emotions~\cite{benlamine2015emotions, 10.1007/978-3-319-58071-5_50, 8631470, li2020emotions, hilton2008emotional}~--~a fact which is insufficiently considered so far.

In this work, we thus investigate for the first time the automatic evaluation of the quality of arguments from a holistic perspective, bringing together various aspects:
First, we evaluate whether \gls{Argument Quality} models can generalize across datasets and domains, which is a crucial feature for deployment in the diverse environments encountered in relevant real-world applications.
Next, we investigate the hypothesis of whether models for related argument mining tasks inherently learn the concept of argument strength without being explicitly trained to do so by evaluating their zero-shot performance for estimating \gls{Argument Quality}.
Finally, we investigate the effect of emotions in arguments:
We present the first dataset for emotions in argumentative texts and demonstrate that emotions can be detected automatically therein, cf. \cref{tab:example}.
The obtained emotion detection models enable us to then provide evidence across all datasets examined that, contrary to the previous belief, emotional argumentation does not significantly influence perceived argument strength.

\begin{table*}
\caption{Two example arguments from the studied datasets with \acrlong{Argument Quality} score and predicted emotion label and model confidence.}
\label{tab:example}
\definecolor{emo}{HTML}{A23114}
\definecolor{nemo}{HTML}{343947}
\begin{tabularx}{\linewidth}{Xcp{2cm}} 
\toprule
\textbf{Topic}: Polygamy Legalization
& \textbf{Score} & \textbf{Emotionality} \\  \midrule
\textcolor{emo}{``Polygamy makes for unhappy relationships and is patriarchal.''} & 0.66 & \textcolor{emo}{emotional} (94.90\%) \\  
\textcolor{nemo}{``Polygamy makes child-raising easier by spreading the needs of children across more people.''} & 0.84 & \textcolor{nemo}{non-emotional} (90.85\%) \\ 
\bottomrule
\end{tabularx}
\end{table*}

In summary, our contributions are as follows:\\
\begin{itemize}
\item We are the first to study the generalization capabilities of \gls{Argument Quality} prediction models across different datasets and \gls{Argument Quality} notions.
\item Since we determine the size of the dataset as one of the decisive performance factors, we further investigate a zero-shot setting of transferring from related Argument Mining tasks.
\item Finally, we elucidate the relation between emotions and \gls{Argument Quality}. To this end, we provide a novel dataset for emotion in argumentative texts and show that these can be predicted on par with human performance. Using this capable emotion detection model, we then show that, in contrast to popular belief, the \gls{Argument Quality} of emotional arguments does \emph{not significantly differ} from non-emotional ones, at least across four different publicly available \gls{Argument Quality} datasets.
\end{itemize}

\section{Related Work}
\glsresetall 
\subsection{Argument Quality}
\gls{Argument Quality}, sometimes also called Argument Strength, is a sub-task in \gls{am} that belongs to the central research topics among argumentation scholars~\cite{walton2008argumentation, toulmin2003uses, van1987fallacies}.
Due to its high subjectivity, there is no single definition of \gls{Argument Quality}.
Therefore, there are various suggestions on different factors that can affect an argument's quality, e.g., the \emph{convincingness} of an argument~\cite{habernal2016makes}.
To the best of our knowledge, we are the first who evaluate how these factors correlate with each other across different corpora.
Furthermore, there are various possibilities to express the strength of an argument.
Some works adopt an absolute continuous score, while others advocate that strength estimation works better in (pairwise) relation to other arguments.

One of the first relatively large corpora was introduced by Swanson et al.~\cite{swanson-etal-2015-argument}.
The \acrshort{SwanRank} corpus contains over 5k arguments, where each argument is labeled with a continuous score that describes the interpretability of an argument in the context of a topic. 
They propose a method using linear regression, ordinary Kriging, and SVMs as regression algorithms to estimate the strength automatically from an input text encoding by handcrafted features.
Other corpora followed and used the relative- and/or absolute \emph{convincingness}~\cite{habernal-gurevych-2016-argument, potash-etal-2019-ranking} as the annotation criterion.
The authors proposed models based on SVMs or BiLSTM combined with GloVe embeddings~\cite{pennington2014glove}.
Gleize et al.~\cite{gleize-etal-2019-convinced} provide a dataset, \emph{IBM-EviConv}, focused on ranking the \emph{evidence} convincingness.
They used a Siamese network based on a BiLSTM with attention and trainable Word2Vec embeddings.
Gretz et al.~\cite{Gretz_Friedman_Cohen-Karlik_Toledo_Lahav_Aharonov_Slonim_2020} and Toledo et al. ~\cite{toledo2019automatic} created their corpora by asking annotators if they would recommend a friend to use the argument in a speech supporting/contesting the topic, regardless of their own opinion.
Both use a fine-tuned BERT~\cite{devlin2019bert} model for the absolute \gls{Argument Quality} regression task.

The shared evaluation practice in the previous works is to evaluate methods on each dataset independently.
Gretz et al.~\cite{Gretz_Friedman_Cohen-Karlik_Toledo_Lahav_Aharonov_Slonim_2020} use the newly introduced dataset for model pre-training but then fine-tune the model on the training part of the dataset used for the evaluation.
This work proposes to advance evaluation and advocate for an accurate \emph{cross-dataset} evaluation without additional fine-tuning on the evaluation dataset to estimate the model's applicability in challenging \emph{real-life} scenarios.

\subsection{Role of Emotions in Argumentation}
The previous works only empirically investigate the role of emotions in argumentation on a small scale.
Wachsmuth et al.~\cite{wachsmuth-etal-2017-computational} created a corpus of 320 arguments, annotated for 15 fine-grained argument dimensions originating from argument theory.
They categorize the quality dimensions into three main quality aspects: \emph{Logical}, \emph{rhetorical}, or \emph{dialectical} quality.
One dimension in rhetorical quality is the \emph{emotional appeal}, defined as:
\emph{``Argumentation makes a successful emotional appeal if it creates emotions in a way that makes the target audience more open to the author’s arguments''}.
The authors did not find any significant correlations to other quality dimensions. 

Benlamine et al.~\cite{benlamine2015emotions, 10.1007/978-3-319-58071-5_50} showed in an experimental setting with 20 participants that the mode of \emph{pathos} represented by emotion is essential in the persuasion process in argumentation. 
Their experiment indicates that \emph{``[the] Pathos strategy is the most effective to use in argumentation and to convince the participants''}.

In both works, the sample size is relatively small (20 participants or 320 arguments).
To better substantiate the considerations, we investigate the influence of emotional appeals regarding \gls{Argument Quality} annotations on more than 40k arguments across four large corpora.


\section{Generalization across Argument Quality Corpora}
\label{sec:generalization}
\glsresetall 
High-level applications such as Argument Retrieval~\cite{wachsmuth2017building, fromm2020diversity, dumani2019systematic, dumani2020framework, stab2018argumentext} and autonomous debating systems~\cite{slonim2021autonomous} require reliable \gls{Argument Quality} models to select strong arguments among the relevant ones.
The research community has identified this gap and proposed and evaluated different automated models for \gls{Argument Quality} estimation~\cite{Gretz_Friedman_Cohen-Karlik_Toledo_Lahav_Aharonov_Slonim_2020,toledo2019automatic,swanson-etal-2015-argument, habernal-gurevych-2016-argument}.
However, \gls{Argument Quality} is often captured differently due to its high subjectivity, e.g., absolutely as a continuous score or relative to other arguments by pairwise comparison.
Consequently, many publications also introduced their corpus with individual annotation schemes capturing different notions of \gls{Argument Quality}.
While they compared multiple models against each other \emph{within} a single corpus, there is a lack of \emph{cross-corpora} empirical evaluations.
Thus, the robustness of predictions across datasets remains largely unexplored, which poses a severe challenge for reliable real-world applications integrating diverse data sources.
To evaluate the generalization capability of \gls{Argument Quality} estimation models, we designed a set of experiments across all four major \gls{Argument Quality} datasets to answer the following research questions:
\begin{itemize}
\item How well do \gls{Argument Quality} models perform across datasets if annotations schema and domain of the arguments do not change?
\item How does the corpora size affect generalization?
\item How well do models generalize across different text domains?
\item How does the \gls{Argument Quality} quality notion affect generalization?
\item Does the \gls{Argument Quality} model become more robust if it is trained with a combined dataset containing data from different domains and labeling assumptions also vary?
\end{itemize}

\subsection{Evaluation Setting}
\label{generalizationDatasets}

\begin{table*}
	
\centering
\caption{
Overview of the different \acrfull{Argument Quality} datasets with their number of arguments, the number of distinct topics, the different source domains, and the \gls{Argument Quality} notion used for annotation.
}
\begin{tabularx}{\linewidth}{X>{\raggedleft\arraybackslash}X>{\raggedleft\arraybackslash}XXX}
\toprule
Name & Sentences & Topics & Domain & Quality notion\\
\midrule
\acrshort{UKPConvArgRank}~\cite{habernal-gurevych-2016-argument} & 1,052 & 32 & Debate Portal  & Convincingness \\
\acrshort{SwanRank}~\cite{swanson-etal-2015-argument} & 5,375 & 4  & Debate Portal & Interpretability \\
\acrshort{IBM-ArgQ}~\cite{toledo2019automatic} & 5,300 & 11 & Crowd Collection & Recommendableness \\
\acrshort{IBM-Rank-30k}~\cite{Gretz_Friedman_Cohen-Karlik_Toledo_Lahav_Aharonov_Slonim_2020} & 30,497 & 71 & Crowd Collection & Recommendableness \\
\bottomrule
\end{tabularx}
\label{tab:generalization}
\end{table*}

We briefly describe the four \gls{Argument Quality} datasets used in our empirical study, which all capture \gls{Argument Quality} on a sentence level.
They are also summarized in \cref{tab:generalization}.
Swanson~et~al.~\cite{swanson-etal-2015-argument} constructed the dataset \texttt{SwanRank} with over 5k arguments whose quality is labeled in the range of $[0, 1]$, where $1$ indicates that an argument can be \emph{easily interpreted}.
Habernal~et~al.~\cite{habernal-gurevych-2016-argument} annotated a large corpus of 16k argument pairs and investigated which argument from the pair is more \emph{convincing}.
Based on the argument pair annotations, they created an argument graph and used PageRank to calculate absolute scores for the individual arguments.
The result is called \texttt{UKPConvArgRank} and contains 1k arguments. 
Gretz~et~al.~\cite{Gretz_Friedman_Cohen-Karlik_Toledo_Lahav_Aharonov_Slonim_2020} and Toledo~et~al.~\cite{toledo2019automatic} created their corpora of 30k and 6.3k arguments by asking annotators if they would \emph{recommend a friend to use the argument} in a speech supporting or contesting the topic regardless of their personal opinion.
Gretz~et~al.~\cite{Gretz_Friedman_Cohen-Karlik_Toledo_Lahav_Aharonov_Slonim_2020} used crowd contributors that presumably better represent the general population, compared to debate club members that annotated in Toledo~et~al.~\cite{toledo2019automatic}.
Furthermore, Gretz~et~al.~\cite{Gretz_Friedman_Cohen-Karlik_Toledo_Lahav_Aharonov_Slonim_2020} also considered the annotators' credibility without removing them entirely from the labeled data, as done in Toledo~et~al.~\cite{toledo2019automatic}.

As some of the corpora did not provide official train-validation-test splits and differed in the number of topics and the formulated task (in-topic vs. cross-topic), we decided to do our own split based on the topics of the arguments.
We perform 10-fold cross-topic cross-validation, where each fold is a 60\%/20\%/20\% train-validation-test split, and we additionally ensure that no topic occurs in more than one split.
By the latter requirement, we ensure an inductive setting where the \gls{Argument Quality} estimation can not rely on similar arguments in the training corpus and therefore provides a more challenging but more realistic task.

\subsection{Model and Training}
\label{generalization_model_training}
Since transfer learning achieves state-of-the-art \gls{am} results on different corpora and tasks~\citep{reimers-etal-2019-classification, DBLP:conf/webi/FrommF019, trautmann2020fine}, we also apply it to our \gls{Argument Quality} estimation task.
We use a bert-base model, pre-trained on masked-language-modeling, and fine-tune it to predict absolute AQ scores on the respective datasets, cf. \cref{generalizationDatasets}.
As an input, we used the arguments from the respective datasets and concatenated the topic information, separated by the BERT specific $[SEP]$ limiter, similar to other work in argument mining \cite{DBLP:conf/webi/FrommF019, reimers-etal-2019-classification, Gretz_Friedman_Cohen-Karlik_Toledo_Lahav_Aharonov_Slonim_2020}.
We concatenate the last four layers of the fine-tuned BERT model output to obtain an embedding vector of the size $4 \cdot 768 = 3,072$.
For the regression task, we stack a \gls{mlp} with two hidden layers, one with 100 neurons and a ReLU activation, followed by the second hidden layer and a sigmoid activation function.
We train the architecture end-to-end, with SGD with a weight decay of 0.35 and a learning rate of $9.1 \cdot 10^{-6}$. The \gls{mlp} uses dropout with a rate of 10\%.

\subsection{Results}
\begin{table*}
\footnotesize
    \centering
    \caption{
    The models are evaluated by the Pearson correlation between ground truth and predicted Argument Quality on the respective test sets.
    The first four rows correspond to models trained on a single dataset, whereas for the last four rows, \emph{all but one} dataset, have been used for training, i.e., following a leave-one-out scheme.
    \textbf{Bold} numbers indicate the best results for each column within the two groups.
    }
    \label{tab:generalization_results}
    \begin{tabularx}{\linewidth}{ll>{\raggedleft\arraybackslash}X|*{4}{>{\raggedleft\arraybackslash}X}}
    \toprule
    \footnotesize
    &&& \multicolumn{4}{c}{Evaluation} \\
    &
    & Size &  \acrshort{UKPConvArgRank} & \acrshort{SwanRank} & \acrshort{IBM-ArgQ} & \acrshort{IBM-Rank-30k} \\
    \midrule
    \multirow{8}{*}{\rotatebox{90}{Training}}
    &\acrshort{UKPConvArgRank} &  1,052 & 19.0\% & 42.5\%  & 15.2\%  & 3.0\%  \\
    &\acrshort{SwanRank} &  5,375 & 18.9\%  & \textbf{47.5\%}  & 17.1\% &  8,0\%  \\
    &\acrshort{IBM-ArgQ} &  5,300 & 23.3\%  & 27.8\%  & 34.2\%  &  38.9\% \\
    &\acrshort{IBM-Rank-30k} & 30,497 &\textbf{26.2\%} & 37.0\% & \textbf{38.3\%}  & \textbf{48.1\%} \\
    \cmidrule(lr){2-7}
    &all except \acrshort{UKPConvArgRank}
    & 41,172 & 23.3\%  & 45.8\%  & 31.6\%  &  46.6\% \\
    &all except \acrshort{SwanRank}
    & 36,849 & \textbf{25.0\%}  & \textbf{49.1\%}  & 35.0\% & 46.6\% \\
    &all except \acrshort{IBM-ArgQ}
    & 36.924 & 23.0\%  & 43.6\% & \textbf{38.4\%}   & \textbf{47.5\%}  \\
    &all except \acrshort{IBM-Rank-30k}
    & 12.224 & 20.4\%  & 42.0\% & 35.0\%  & 46.5\%  \\
    \bottomrule
    \end{tabularx}
\end{table*}

\cref{tab:generalization_results} summarizes our results.
We report the Pearson correlation score between the predicted- and ground-truth absolute \gls{Argument Quality} evaluated on a hold-out test set.

\subsubsection{Evaluation on Similar Datasets and Importance of Training Set Size}
First, we evaluate the performance of the model on similar datasets and the dependency on the size of the training dataset.
We can observe that models perform very well on other datasets from the same domain labeled with a similar quality notion, i.e., \acrshort{IBM-ArgQ} and \acrshort{IBM-Rank-30k} datasets.
Furthermore, we can notice that the size of the dataset is crucial for performance:
a model trained on the largest \acrshort{IBM-Rank-30k} dataset achieves the best score also on \acrshort{IBM-ArgQ}.
This insight gives us a solid foundation for the next steps.
\subsubsection{Generalization Across Domains and Quality Notions}
Next, we investigate whether a transfer across domains is possible.
To this end, we train on one dataset and evaluate on a different one.
Recall that the four datasets cover two different domains:
the sentences from \acrshort{UKPConvArgRank} and \acrshort{SwanRank} have been extracted from debate portals, while \acrshort{IBM-Rank-30k} and \acrshort{IBM-ArgQ} have been collected from the crowd.

Compared to in-domain generalization, we observe a considerably worse generalization between domains:
For example, trained on the crowd dataset \acrshort{IBM-ArgQ}, we can achieve a correlation of 38.9\% on the crowd dataset \acrshort{IBM-Rank-30k}, while training on the debate datasets \acrshort{SwanRank} and \acrshort{UKPConvArgRank} results in negligibly low correlations of 8\% and 3\%, respectively.
Conversely, when evaluated on the debate portal dataset \acrshort{SwanRank}, we obtain a correlation of 42.5\% when using a model trained on the other debate portal dataset \acrshort{UKPConvArgRank}, while the crowd collected datasets \acrshort{IBM-ArgQ} and \acrshort{IBM-Rank-30k} only achieves 27.8\% and 37.0\%, respectively.
The smaller difference compared to the first comparison can be explained by the larger size of the training datasets.

Surprisingly, we observe a completely different picture for generalization across quality notions. We see only a moderate drop in performance for a fixed domain but a different quality notion. For instance, the model trained on \acrshort{SwanRank} performs relatively well on the \acrshort{UKPConvArgRank} dataset. Vice-versa, we observe a more considerable performance drop, which can be explained by the smaller size of the \acrshort{UKPConvArgRank} dataset.

\subsubsection{Multi-Domain and Multi-Quality Notion Training}
To investigate whether a single model can grasp various dimensions of quality and work on arguments from various domains, we designed another set of ``leave-one-out'' experiments.
We train on the training sentences of all but one \gls{Argument Quality} corpus and evaluate the performance on all test sets.
The entries on the diagonal thus show how well the models perform when evaluated on an unseen corpus.

For evaluation on the unseen \acrshort{IBM-Rank-30k} dataset after training on the remaining ones, we can obtain a correlation of 46.5\%, which nearly reaches the correlation of 48.1\% we obtained when training and evaluating on \acrshort{IBM-Rank-30k}.
For \acrshort{SwanRank}, \acrshort{IBM-ArgQ} and \acrshort{UKPConvArgRank}, we can even surpass the correlation on the respective test set by training on all other training sets instead of the one from the respective corpus. 

\subsubsection{Cross-Corpora Generalization Conclusion}
To summarize, we conclude that, in general, the available datasets and models for \gls{Argument Quality} are reliable, and the models can grasp the concepts automatically.
Our most important insight is that \gls{Argument Quality} notions do not contradict each other, and a single model can estimate the \gls{Argument Quality} of text from different domains. 
Therefore, the practical recommendation for real-life application is to combine all available datasets across different domains and \gls{Argument Quality} notions.

\section{Zero-Shot-Learning in Argument Mining}
\label{sec:task-generalization}
\glsresetall 
In this section, we investigate whether explicit \gls{Argument Quality} corpora are a necessity, or whether the task of \gls{Argument Quality} can also be solved by transferring from other related argument mining tasks such as \gls{Argument Identification} or \gls{Evidence Detection},
In contrast to the relatively new task of automatic \gls{Argument Quality} estimation, other \gls{am} tasks already offer a broad range of large datasets that cover different domains and annotation schemes.
Moreover, the agreement between the annotators is higher on the other tasks, as \gls{Argument Quality} is highly subjective~\cite{swanson-etal-2015-argument, Gretz_Friedman_Cohen-Karlik_Toledo_Lahav_Aharonov_Slonim_2020, toledo2019automatic,habernal-gurevych-2016-argument, stab-etal-2018-cross}.
Therefore, a successful transfer from related tasks to the target task of \gls{Argument Quality} would represent a significant advance in the field.
To this end, we investigate the zero-shot capability of \gls{am} models across different corpora \emph{and} different \gls{am} tasks.
To the best of our knowledge, we are the first to compare \gls{am} task similarity by providing a first study on how individual tasks can benefit from each other.

In particular, we aim to answer the following guiding research questions:
\begin{itemize}
    \item Can we achieve satisfactory performance by zero-shot transfer from related \gls{am} tasks, i.e., without fine-tuning the respective task?
    \item Is there a difference in transferring from different tasks, i.e., is one task more suited than the other?
\end{itemize}
While not a primary focus of this work, for completeness, we also provide experimental results for the reverse direction of transferring \emph{from} \gls{Argument Quality} estimation \emph{to} the other tasks.

\subsection{Datasets}
\begin{table*}
\small
\centering
\caption{
Overview of the different \acrfull{am} datasets, we used for the zero-shot experiments, with their size in terms of the number of sentences, the number of covered topics, the source domain and the \gls{am} task.
}
\label{tab:zero-shot}
\begin{tabularx}{\linewidth}{l*{2}{>{\raggedleft\arraybackslash}X}Xl}
\toprule
Name & Sentences & Topics & Domain & Task\\
\midrule
\acrshort{IBM-Rank-30k}~\cite{Gretz_Friedman_Cohen-Karlik_Toledo_Lahav_Aharonov_Slonim_2020} & 30,497 & 71 & Crowd Collection & \acrfull{Argument Quality} \\
\acrshort{UKP-Sentential}~\cite{stab-etal-2018-cross} & 25,492 & 8 & Web Documents & \acrfull{Argument Identification} \\
\acrshort{IBM-Evidence}~\cite{ein2020corpus} & 29,429 & 221 & Wikipedia & \acrfull{Evidence Detection}\\
\bottomrule
\end{tabularx}
\end{table*}

\cref{tab:zero-shot} provides an overview of the different \gls{am} corpora we used in our experiments, covering three different \gls{am} tasks.
\acrshort{UKP-Sentential}~\cite{stab-etal-2018-cross} contains over 25k arguments distributed across eight controversial topics.
It is annotated for \gls{Argument Identification}, where each argument is labeled as either \emph{argumentative} or \emph{non-argumentative} in the context of a topic.
The \acrshort{IBM-Evidence}~\cite{ein2020corpus} corpus includes nearly 30k sentences from Wikipedia articles.
All sentences are annotated with a score in the range of $[0, 1]$, denoting the confidence that the sentence is evidence (either expert or study evidence) to the article's topic.
\acrshort{IBM-Rank-30k}~\cite{Gretz_Friedman_Cohen-Karlik_Toledo_Lahav_Aharonov_Slonim_2020} is the largest of the four \gls{Argument Quality} datasets, which has also been used in the previous \cref{sec:generalization}.
The corpus' annotation is in the range of $[0, 1]$, where $1$ indicates a strong argument and a score of $0$ indicates a weak argument.
We split all three datasets into train, validation, and test sets (70\%/10\%/20\%).
Similar to \cref{generalizationDatasets}, we designed the splits such that no topic in the training set also occurs in the test set, which is often called the "cross-topic" scenario in \gls{am} and corresponds to a more interesting, but also more challenging task, which requires a sufficient degree of generalization to unseen topics.

\subsection{Evaluation Setting}
We use a standard BERT large model~\cite{devlin2019bert} pre-trained on the masked-language-modeling task to evaluate the zero-shot generalization capability.
As an input for the fine-tuning, we use the sentences from the respective datasets and concatenate the topic information, separated by the BERT specific \texttt{[SEP]} limiter, similar to \cref{generalization_model_training}.
We develop three different zero-shot evaluation strategies for the different transfer settings:
\begin{itemize}
    \item \textbf{\gls{Argument Identification} $\rightarrow$ Regression Tasks}:
    We use the BERT encoder output as input to a linear layer with dropout that predicts the classes.
    Cross-entropy serves as training loss.
    The probabilities between $0$ and $1$ indicate if a sentence is argumentative or not.
    The predicted probability of the positive class, i.e., whether it is argumentative, is then directly used as a score for \gls{Evidence Detection} and \gls{Argument Quality} on the respective corpora.
    We use Spearman rank-correlation instead of Pearson correlation as an evaluation measure to account for the difference in scale.
    
    \item \textbf{Regression Tasks $\rightarrow$ \gls{Argument Identification}}:
    \gls{Evidence Detection} and \gls{Argument Quality} use the BERT representations in a single hidden layer that scores the sentences according to their absolute quality or the probability of containing evidence.
    Since we train on regression tasks, we use the Mean Squared Error loss during training.
    We then apply the trained models to \gls{Argument Identification}.
    We select an optimal decision threshold $\alpha$ among all possible thresholds on \acrshort{UKP-Sentential}'s validation set according to Macro $F_1$.
    This model is then evaluated on the \acrshort{UKP-Sentential} test set.
    
    \item \textbf{Regression Task $\leftrightarrow$ Regression Task}:
    For the evaluation between two regressions models, we calculate the Spearman correlation coefficient directly on their respective outputs.

\end{itemize}

\subsection{Results}


\begin{table*}
		
    \caption{Zero-Shot performance of the \acrlong{am} models. The evaluation measure is Macro F$_{1}$ for \acrfull{Argument Identification}, and the Spearman correlation for \acrfull{Evidence Detection} and \acrfull{Argument Quality}.
    }
    \label{tab:zero_results}
    \begin{tabularx}{\linewidth}{l>{\raggedleft\arraybackslash}X|*{2}{>{\raggedleft\arraybackslash}X}}
    \toprule
    Train & \multicolumn{3}{c}{Evaluation} \\
    & \multicolumn{1}{c}{\acrshort{Argument Identification}}  &
    \multicolumn{1}{c}{\acrshort{Evidence Detection}} & \multicolumn{1}{c}{\acrshort{Argument Quality}} \\
    \midrule
    \acrshort{Argument Identification}  & $73.51\% \pm 3.37\%$
    & $55.53\% \pm 1.17\%$  & $27.49\% \pm 1.54\%$ \\
    \acrshort{Evidence Detection} & $75.16\% \pm 0.71\%$
    & $77.90\% \pm 0.24\%$ & $28.66\% \pm 0.92\%$   \\
    \acrshort{Argument Quality}  & $71.27\%\pm 0.74\%$
    & $43.50\% \pm 3.10\%$ & $47.45\% \pm 1.16\%$  \\
    \midrule
    Metric: & 
    \multicolumn{1}{c}{Macro $F_{1}$} & \multicolumn{1}{c}{$\rho$} & \multicolumn{1}{c}{$\rho$}\\
    \bottomrule
    \end{tabularx}
\end{table*}
\cref{tab:zero_results} shows the results from our experiments.
We train three models with different random seeds for each training task and report the mean and standard deviation of evaluation on the different tasks.

We generally observe, unsurprisingly, that training on the same task as evaluating yields the best results with Spearman correlations of $\approx 77.90\%$ for \gls{Evidence Detection} $\to$ \gls{Evidence Detection} and $\approx 47.45\%$ for \gls{Argument Quality} $\to$ \gls{Argument Quality}.

A notable exception is \gls{Argument Identification}, where a model trained on \gls{Evidence Detection} achieves $\approx 75.17\%$ Macro $F_1$ and thus can slightly surpass the performance of a model directly trained on \gls{Argument Identification} of $\approx 73.53\%$, although within the range of one standard deviation.
Exceeding the in-task performance is a strong result, as the model has never explicitly been trained for the task.
We generally observe almost perfect zero-shot transfer towards \gls{Argument Identification}, as also the model trained on \gls{Argument Quality} achieves a performance of $\approx 71.27\%$, which is only 2\% points behind the $\approx 73.53\%$ from \gls{Argument Identification} to \gls{Argument Identification}.
Thus, models capable of predicting whether a sentence provides evidence (\gls{Evidence Detection}) or capable of predicting the \gls{Argument Quality} of an argument, inherently learn concepts that enable the detection of whether a sentence is argumentative or not (\gls{Argument Identification}).
To further give context to the zero-shot performance, the BiCLSTM approach trained on the \gls{Argument Identification} task from~\cite{stab-etal-2018-cross} obtained a Macro F$_{1}$ of 64.14\%, i.e., worse results than the zero-shot transfer despite explicitly being trained on the task, which underlines the remarkable zero-shot performance, and may indicate that \gls{Argument Identification} is a simpler task than the other two, \gls{Evidence Detection} and \gls{Argument Quality}.

For \gls{Evidence Detection}, we achieve the best performance of $\approx 77.90\%$ Spearman correlation by directly training on this task.
The model trained on \gls{Argument Identification} obtains the closest zero-shot transfer result with a rank correlation of $\approx 53.80\%$, which still represents a considerable correlation, despite being $\approx 24\%$ points behind.
The model trained for \gls{Argument Quality} shows the worst transfer from the studied tasks with a correlation of $\approx 43.51\%$.
Overall, we note that the challenging zero-shot transfer is still possible with an acceptable loss in performance.
Models trained on detecting whether a sentence is argumentative or not (\gls{Argument Identification}) transfer better than those trained for predicting the argumentative strength of a sentence \gls{Argument Quality} to the target task of predicting the confidence whether a sentence provides evidence (\gls{Evidence Detection}).

For \gls{Argument Quality}, the main focus of our paper, we achieve the best performance of $\approx 47.45\%$ Spearman correlation by directly training on this task.
When transferring from related \gls{am} tasks in a zero-shot setting, we have to tolerate decreases in performance to $\approx28.66\%$ for transfer from \gls{Evidence Detection}, and $\approx25.72\%$ for transfer from \gls{Argument Identification}, respectively.
Thus, models capable of detecting whether a sentence is argumentative (\gls{Argument Identification}) are slightly less well applicable to predicting the sentence's argumentative strength than the models for predicting a level of supporting evidence (\gls{Evidence Detection}).
One factor here may be that \gls{Evidence Detection} is also a regression task as opposed to the classification task of \gls{Argument Identification}.

To summarize, the results suggest that the tasks of \gls{Argument Identification}, i.e., classifying whether a sentence is argumentative, and \gls{Evidence Detection}, i.e., predicting a numeric level of supporting evidence, are closer to each other than to the more difficult task of assessing the argumentative strength, as witnessed by worse zero-shot transfer results from and to \gls{Argument Quality}.
Nevertheless, in principle, a transfer in the highly challenging zero-shot setting is possible; for closer related tasks, it can even lead to similar scores as training directly on the target task.

\subsection{Multi-Task Learning for Argument Quality}
\begin{table*}
	
    \caption{
    Performance of multi-task models trained on different \acrlong{am} task combinations, including \acrfull{Argument Identification} and \acrfull{Evidence Detection}.
    The performance is measured by Macro F$_{1}$ for \gls{Argument Identification}, and the Spearman correlation for \gls{Evidence Detection} and \gls{Argument Quality}.
    }
    \label{tab:multi-task}
    \begin{tabularx}{\linewidth}{l*{3}{>{\raggedleft\arraybackslash}X}}
    \toprule
    Train & \multicolumn{3}{c}{Evaluation} \\
    & \multicolumn{1}{c}{\acrshort{Argument Identification}}  &
    \multicolumn{1}{c}{\acrshort{Evidence Detection}} & \multicolumn{1}{c}{\acrshort{Argument Quality}} \\
    \midrule
    \acrshort{Argument Quality} & - & - & $47.45\% \pm 1.16\%$ \\
    \acrshort{Argument Quality}/\gls{Argument Identification} & $80.07\%\pm 1.16\%$  & - & $47.46\% \pm 0.58 \%$ \\
    \acrshort{Argument Quality}/\gls{Evidence Detection} & - & $78.07\% \pm 0.45\%$ & $46.84\% \pm 0.25\%$  \\
    \acrshort{Argument Quality}/\gls{Argument Identification}/\gls{Evidence Detection} & $78.91\% \pm 3.17\%$ & $ 78.40\% \pm 0.03\%$  & $48.39\% \pm 1.12\%$ \\
    \midrule
    Metric: & 
    \multicolumn{1}{c}{Macro $F_{1}$} & \multicolumn{1}{c}{$\rho$} & \multicolumn{1}{c}{$\rho$}\\
    \bottomrule
    \end{tabularx}
\end{table*}
As shown in the last section, the \gls{am} tasks are sufficiently close to each other to enable successful zero-shot transfer.
An interesting question that arises from this observation is whether the performance in \gls{Argument Quality} estimation further improves by multi-task learning.
To this end, we developed a multi-task model that involves a shared BERT encoder and separated linear layers for the respective tasks.
We trained the architecture with weighted loss functions, ensuring that each task is weighted equally.
Our results are shown in \cref{tab:multi-task}.
Focusing on the right-most column first, we can see that the performance in terms of Spearman correlation only marginally improves by multi-task learning.
A possible explanation is here that we already observed that the other two tasks are seemingly less challenging and more closely related to each other than to \gls{Argument Quality}.
As additional supporting evidence, \gls{Evidence Detection} slightly and \gls{Argument Identification} considerably benefits from multi-task learning with \gls{Argument Quality}.

\section{Emotion Detection}
\label{sec:emotionDetection}
\glsresetall 
Most work in \gls{am} focuses on the \emph{logos} mode of persuasion, i.e., whether arguments are logically plausible.
Nevertheless, recent studies support that the mode of \emph{pathos} represented by emotions is essential in the persuasion process~\cite{benlamine2015emotions, 10.1007/978-3-319-58071-5_50, 8631470, li2020emotions, hilton2008emotional}.
Those studies have in common that they relied on relatively small sample sizes.
In the following, we thus evaluate the hypothesis that the \acrshort{Argument Quality} scores in the publicly available \gls{Argument Quality} datasets Swanson~\cite{swanson-etal-2015-argument}, UKP~\cite{habernal-gurevych-2016-argument}, Gretz~\cite{Gretz_Friedman_Cohen-Karlik_Toledo_Lahav_Aharonov_Slonim_2020}, and Toledo~\cite{toledo2019automatic}, are influenced by appealing to the emotions of the annotators.

\gls{Argument Quality} datasets do not provide emotion labels, and therefore, we first need a reliable and scalable method to estimate the level of emotionality in the arguments. As to the best of our knowledge, there is no previous work on automatic emotion detection in arguments, we investigate various approaches from the very simple baselines to complex multi-step transfer learning models.
For the evaluation and comparison of different methods, we create a novel argument dataset \acrshort{EmoArg523}, where for each argument, we manually annotate emotionality.

After the reliable emotion detection model is available, we apply it on the unlabeled arguments from the four \gls{Argument Quality} corpora to obtain proxy emotion labels.
We then use these proxy labels to investigate the relation between emotions and argument \gls{Argument Quality} at a large scale.

In particular, we address the following research questions:
\begin{itemize}
    \item Can we automatically detect emotions in argumentative texts from different domains?
    \item Can we substantiate the hypothesis that arguments arousing emotions are perceived stronger?
\end{itemize}

\subsection{Datasets}
\label{emotiondatasets}

\subsubsection{Novel Emotional Argumentation Dataset}
For the evaluation of our \acrshort{BertBaseMT} model, we sample 150 arguments from each of the four \gls{Argument Quality} corpora (\acrshort{IBM-Rank-30k} , \acrshort{IBM-ArgQ}, \acrshort{SwanRank}, \acrshort{UKPConvArgRank}), i.e., 600 in total.
These arguments are manually labeled by six independent annotators.
The arguments were labeled based on the annotation guidelines as \emph{emotional}, when the arguments contained pathos rhetoric, or \emph{non-emotional} when the persuasion process in the argument was driven by evidence or logical rhetoric.
The annotator agreement calculated via Krippendorff's Alpha~\cite{krippendorff2016reliability} is 31.28\%.
Note that because of the subjectivity of the task, such an agreement is acceptable; for comparison, e.g., Wachsmuth et al.~\cite{wachsmuth-etal-2017-computational} achieved an Alpha of 26\% for the quality dimension ``Emotional appeal``.
After the agreement calculation, we removed the 77 sentences (12.8\%) without a majority between the six annotators.
The resulting dataset comprises 225 emotional (43.02\%) and 298 (56.98\%) non-emotional arguments and is referred to as \acrshort{labeledArgEmotions}.
We split the dataset into train-, validation-, and test-set (60\%/10\%/30\%).

\subsubsection{General Emotion Detection Dataset}
\label{subsubsec:gedd}
\begin{table*}
	
\centering
\caption{Overview of the different emotion detection datasets from heterogeneous text domains used for the behavioral fine-tuning of \acrshort{BertBaseMT}.}
\label{tab:emotiondatasets}
\begin{tabularx}{\linewidth}{lrX}
\toprule
Name & Sentences & Domain\\
\midrule
\acrshort{Alm}~\cite{Alm2009AffectIT} & 15,036  & Childrens' stories\\
\acrshort{ISEAR}~\cite{Scherer1994EvidenceFU} &  7,666  & Reactions and emotion antecedents\\
\acrshort{SemEval-2007}~\cite{10.5555/1621474.1621487}& 1,250 & News headlines\\
\acrshort{SemEval-2018}~\cite{mohammad-etal-2018-semeval} & 9,625 & Tweets\\
\acrshort{SemEval-2019}~\cite{chatterjee-etal-2019-semeval} & 14,335  & Dialogues\\
\acrshort{Neviarouskaya 2010}~\cite{neviarouskaya-etal-2010-recognition} & 1,000  & Stories \\
\acrshort{Neviarouskaya 2011}~\cite{neviarouskaya_prendinger_ishizuka_2011} & 700 & Diary-like-blogs \\
\bottomrule
\end{tabularx}
\end{table*}
Although it is not clear a priori that emotion detection transfers well from other domains to the domain of argumentation, we hypothesize that the model can benefit from existing datasets. 
Motivated by our results for \gls{Argument Quality} detection, where the model trained on a joined dataset demonstrated very robust performance, we combine seven emotion datasets~\cite{Alm2009AffectIT, Scherer1994EvidenceFU, 10.5555/1621474.1621487, mohammad-etal-2018-semeval, chatterjee-etal-2019-semeval, neviarouskaya-etal-2010-recognition, neviarouskaya_prendinger_ishizuka_2011}, c.f., \cref{tab:emotiondatasets}. 
The emotion datasets came with different classes of emotions.
Thus, we unified these different label formats by assigning the existing labels for emotions, such as \textit{happy}, \textit{sad} or \textit{fear}, or \textit{neutral}, to a binary label of either \textit{emotional} or \textit{non-emotional}.
The seven datasets were then split individually into train-, validation-, and test-set and combined to a large heterogeneous emotion corpus.

\subsection{Models \& Baselines}
Since transfer learning achieves state-of-the-art results for AM on different corpora and tasks~\citep{reimers-etal-2019-classification, DBLP:conf/webi/FrommF019, trautmann2020fine}, we also apply it for the task of emotion detection.
We employ a transformer~\cite{vaswani2017attention} based BERT model~\cite{devlin2019bert} with fine-tuning on different datasets.
As a regularization technique to avoid over-fitting, early stopping is used on the validation cross-entropy loss, with a patience value of three epochs.
We include the following model variants and baselines in our evaluation:

\begin{description}
    \item[\acrshort{BaseMaj}]
    The majority baseline labels the arguments with the most frequent class based on our \acrshort{labeledArgEmotions} corpus, which is \emph{non-emotional} (57.55\%).
    \item[\acrshort{BasePro}]
    The pronouns baseline labels the arguments as emotional, which contain at least one of the personal pronouns "I", "you" or "me".
    \item[\acrshort{BaseNRC}]
    The NRC baseline labels the arguments which contain at least \emph{one} unigram contained in the NRC Emotion Lexicon~\cite{mohammad2013crowdsourcing}.
    \item [\acrshort{BertBaseMT}] 
    To assess the generalization, we evaluate the zero-shot performance of a BERT model fine-tuned on the heterogeneous emotion corpora, cf. see \cref{subsubsec:gedd}.
    The combined emotion corpora incorporate multiple domains found on the internet, and therefore, the resulting model is supposed to be universally applicable.
    \item[\acrshort{BertBaseAM}] Bert-Base fine-tune it on 339 sentences of our annotated argument emotion dataset.
    \item [\acrshort{BertBaseMTAM}] 
    It is the same as \acrshort{BertBaseMTAM}, but we also fine-tune it on 339 sentences of our annotated argument emotion dataset.
    We hypothesize that with the two-step transfer learning approach, the model first learns a general concept of emotions and then can focus on the target argument domain.
    \item[\acrshort{HP}] 
    An interesting experiment for assessing the applicability of the proposed solution is the comparison with the human performance on the task.
    To compute the human performance, we evaluate each annotator against the majority label of the remaining annotators using the Macro $F_{1}$ score.
\end{description}

\subsection{Results}
\subsubsection{Emotion Detection}

\begin{table*}
    \centering
    \caption{%
        Overview of the emotion detection results for different model variants on the annotated \acrlong{Argument Quality} dataset, \acrlong{labeledArgEmotions}.
        We show results in terms of Macro F$_{1}$ for different BERT model variants, as well as the three baselines in addition to a human performance estimate.
        For those models which do not make use of the labels on \acrlong{labeledArgEmotions}, we also report the performance across all labeled arguments.
        In \textbf{bold} font, we highlight the best performance inside one group.
    }
    \label{tab:emotion_detection}
    \begin{tabularx}{\linewidth}{Xrr}
    \toprule
    Method & \multicolumn{2}{c}{Split} \\
    & train+val+test & test \\
    \midrule
    \acrshort{BaseMaj} & 36.3\% & 36.4\%  \\
    \acrshort{BasePro} & \textbf{63.0\%}  & \textbf{59.7\%} \\
    \acrshort{BaseNRC} & 52.3\%  & 50.3\% \\
    \midrule
    \acrshort{BertBaseMT} & $\mathbf{67.1\% \pm 3.0\%}$  & $65.9\% \pm 5.3\%$\\
    \acrshort{BertBaseAM} & - & $73.2\% \pm 1.8\%$  \\
    \acrshort{BertBaseMTAM} & - &$\mathbf{74.6\% \pm 1.7\%}$  \\
    \midrule
    \acrshort{HP} & $82.1\% \pm 4.0\%$ & $80.9\% \pm 4.0\%$ \\
    \bottomrule
    \end{tabularx}
\end{table*}

We present the results in terms of Macro $F_1$ on the novel dataset \acrshort{labeledArgEmotions} for emotion detection in argumentative texts in \cref{tab:emotion_detection}.
Despite its simplicity, the strongest baseline with a Macro $F_1$ score of $59.7\%$ is the \gls{BasePro},
\acrshort{BertBaseMT} achieves a Macro $F_1$ of $\approx 67.1\%$, which highlights that \emph{domain adoption} from the source - the heterogeneous emotion detection datasets - to the target domain of arguments \emph{is possible}.
The best emotion detection model, \acrshort{BertBaseMTAM}, which used behavioral fine-tuning on the emotion dataset, followed by a second fine-tuning on the dataset of emotion-annotated arguments, achieves a Macro $F_1$ score of $\approx 74.6\%$, only a few points below the human performance estimate of $\approx 80.9\%$.
For most models, we also observe a slight decrease in performance between the test part and the full \gls{labeledArgEmotions}; we attribute this to a slight distribution shift where the test part seems to contain slightly more arguments with difficult to detect emotions.

\subsubsection{The Effect of Emotions on Argument Quality}
\begin{figure*}
\includegraphics[width=\linewidth]{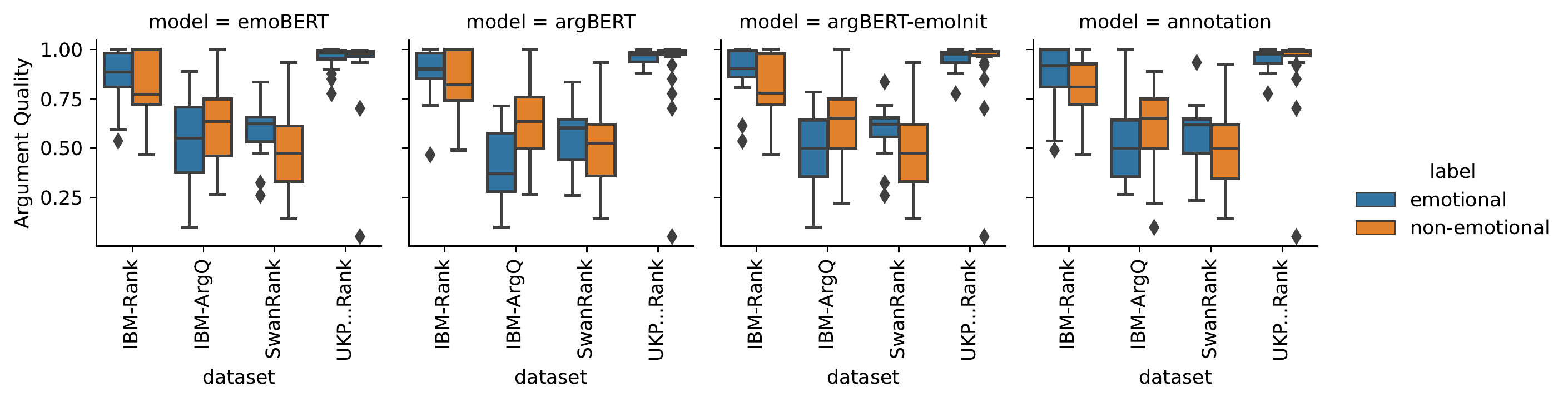}
\caption{
Relation between predicted / annotated emotionality and \acrlong{Argument Quality} grouped by dataset for the three different models and the ground truth (on the right-most panel; denoted by \emph{annotation}).
}
\label{fig:emotion-vs-strength-small}
\end{figure*}
We start by analyzing the relation of emotionally appealing texts and \gls{Argument Quality} on the relatively small test part of the novel annotated dataset, \gls{labeledArgEmotions}.
\cref{fig:emotion-vs-strength-small} shows the distribution of \gls{Argument Quality} for emotional vs. non-emotional arguments based on the three different emotion detection models and the ground truth annotation grouped by dataset.
Except for \acrshort{IBM-ArgQ}, we observe the mean \gls{Argument Quality} of emotional arguments to be higher than those of non-emotional arguments. A possible explanation is that, in contrast to the other datasets that used crowd workers, the annotation on \acrshort{IBM-ArgQ} was created by debate club members, who may have been trained to judge explicitly not considering an emotional appeal.
However, partially due to the small sample size, the differences are insignificant ($p>0.01$) according to Welsh's unequal variance t-test with Fischer adjustments.

\begin{figure*}
\includegraphics[width=\textwidth]{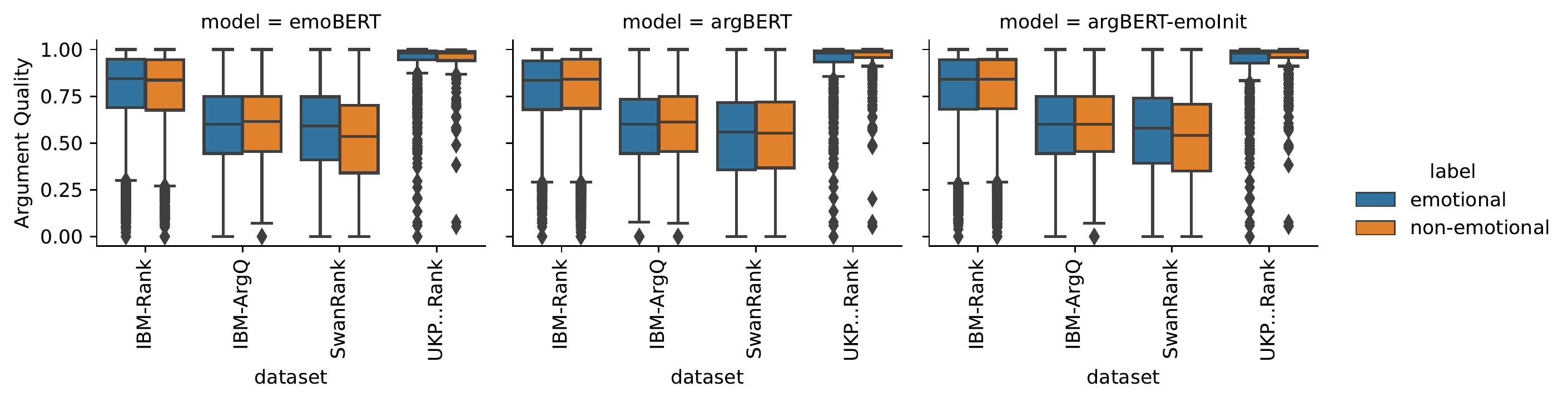}
\caption{
Comparison of the \acrfull{Argument Quality} of the remaining unlabeled $>40,000$ arguments grouped by predicted emotionality across the four datasets.
}
\label{figure:emotion-vs-strength-large}
\end{figure*}
Next, we utilize the trained emotion detection models to extend the analyses from the 157 test sentences in \gls{labeledArgEmotions} to the remaining 41,905 from the combined \gls{Argument Quality} corpora.
While we are now restricted to predicted emotionality only instead of human annotations, we reviewed its quality in the previous section and found it sufficient.
\cref{figure:emotion-vs-strength-large} shows the distribution of \gls{Argument Quality} grouped by predicted appeal to emotion for all three models and four datasets.

For \acrshort{SwanRank}, emotional arguments receive slightly larger quality scores.
While this is consistent across all models, it is clearly visible for \acrshort{BertBaseMT} and the most reliable emotion prediction model, \acrshort{BertBaseMTAM}.
We attribute this to the covered topics of Gun Control, Gay Marriage, Death Penalty, and Evolution, which are areas with emotional discussions.
On \acrshort{IBM-ArgQ}, the differences are smaller but consistent across all models, with a slight tendency towards non-emotional arguments being perceived stronger.
A possible explanation can be the annotation process, where debate club members served as annotators, which may be taught towards looking at logical arguments without letting emotions affect their view.
The other two datasets do not show a consistent nor noticeable difference in the distributions.
Overall, we cannot observe a clear relation between emotional argumentation and perceived strength on a large scale, challenging existing views.
With our new dataset for emotional argumentation and the proxy models capable of predicting emotional argumentation, we hope to enable social sciences to study the deeper reasons behind this in the future.

\section{Conclusion}
\glsresetall 
We see this work as a fundamental step towards a holistic view of \gls{Argument Quality}:
We showed that for good generalization across individual \gls{Argument Quality} corpora, a match between the source and target domain of the arguments is essential.
In contrast, diversity in \gls{Argument Quality} notions does not hinder but rather enriches the generalization capability.
The target domain has a minor impact with sufficient broad coverage of different domains and adequate size.
This insight is directly actionable for practical applications:
The benefits of different \gls{Argument Quality} notions permit direct integration of different data sources, which is a prerequisite for dealing with the inputs from diverse domains encountered, e.g., by general-purpose argument retrieval engines.

Moreover, we could elucidate \gls{Argument Quality}'s relation to other \gls{am} tasks, such as  \gls{Evidence Detection} and \gls{Argument Identification}.
Our zero-shot transfer experiments demonstrated that the concepts learned for one of the tasks are sufficient to solve the other to some degree without explicitly being trained for it.
By comparing the achieved results, we conclude that \gls{Argument Identification} and \gls{Evidence Detection} are more closely related to each other than to \gls{Argument Quality}, and per se also easier to transfer to it.
The multi-task experiment further emphasized this, where \gls{Argument Quality} could gain less from the other tasks than vice-versa.
Thus, an important open question is how to enable better successful transfer towards \gls{Argument Quality}, and also extending beyond the three tasks we studied in this work.

Finally, we provide the community with a new corpus that consists of \gls{Argument Quality} \emph{and} emotion annotations.
In contrast to some results from social science research, our extensive empirical evaluation across a large number of argumentative sentences found overall only a limited influence of emotional appeal on the \gls{Argument Quality} scores.
A deeper analysis of these surprising results' (social) determinants is an important future work direction.
Besides the well-studied \emph{logos} dimension of logical plausibility and the \emph{pathos} dimension investigated in this work, the third remaining dimension from classical argumentation theory is \emph{ethos}, which did not receive sufficient attention by the \gls{am} community so far.
Existing smaller datasets \cite{koszowy2022theory, duthie2016mining} invite to visit these uncharted territories, e.g., by studying argument strength in context to the provenance of an argument.

\bibliography{anthology,custom}
\bibliographystyle{acl_natbib}

\appendix

\clearpage

\section{Computing \& Software Infrastructure}
\label{app:infrastructure}
All experiments were conducted on a Ubuntu 20.04 system with an AMD Ryzen Processor with 32 CPU-Cores and 126 GB memory. We further used Python 3.7, PyTorch 1.4, and the Huggingface-Transformer library (4.15.0). For the experiments in Chapter 3, we used four NVIDIA RTX 2080 TI GPU with 11 GB memory. The models in Chapter 4 and 5 were trained on a single NVIDIA Tesla V100. The default parameters from the Huggingface-Transformer library \footnote{\url{https://huggingface.co/docs/transformers/master/en/main_classes/trainer\#transformers.TrainingArguments}} were used for all hyperparameters not specified in the following sections.

\section{GENERALIZATION ACROSS ARGUMENT QUALITY CORPORA}
In \cref{sec:generalization}, we trained bert-base-uncased models with a batch size of 64. The learning rate was set to $9.1 \cdot 10^{-6}$. A weight decay of $0.31$ was used. We calculated the 95th percentile based on the four AQ validation sets and truncated longer sentences to that length. We used a dropout rate of $0.1$ for the dropout layer in the \textbf{\gls{Argument Identification} $\rightarrow$ Regression Tasks} setting. The losses in the multi-dataset setting were equally weighted for each of the four datasets. We used early stopping on the validation MSE loss, with a patience value of five epochs, as a regularization technique to avoid over-fitting.

\section{ZERO-SHOT-LEARNING IN ARGUMENT MINING}
For \cref{sec:task-generalization}, we trained bert-large-uncased architectures with a batch size of 64. The learning rate was set to $1 \cdot 10^{-5}$, and for the first 0.1 epochs, a warm-up period is used. We opt for evaluations every 0.1 epochs in our training configuration, resulting in 10 evaluations per epoch. Our train/validation/test split is based on a reasonably standard 70\%/10\%/20\% split. Furthermore, we calculate the 99th percentile of the max length of all sentences inside the validation split, and truncate them to that length. This further decreases the required learning time, due to a reduced input dimension without losing significant information. The losses in the multi-dataset and multi-task setting were equally weighted for each of the three argument mining datasets. Finally, to further reduce variance in training, we use three seeds for our experiments and calculate the mean and standard deviation for all of our results.

\section{EMOTION DETECTION}
For \cref{sec:emotionDetection}, we trained bert-base-cased architectures with a batch size of 32. The learning rate was set to $5 \cdot 10^{-5}$ A weight decay of $0.1$ was used.  We opt for evaluations every 0.25 epochs in our training configuration, resulting in 4 evaluations per epoch. The annotators used the Inception Annotation Framework~\footnote{\url{https://inception-project.github.io/releases/22.3/docs/user-guide.html}} for the labeling of the arguments. The annotated dataset is split into train/validation/test (60\%/10\%/30\%). Furthermore, we calculate the 99.5th percentile of the max length of all sentences in the validation split, and truncate all sentences to that length. 

\end{document}